\documentclass[conference]{IEEEtran}
\usepackage{cite}
\usepackage{amsmath,amssymb,amsfonts}
\usepackage{graphicx}
\usepackage{textcomp}
\usepackage{xcolor}
\usepackage{url}

\begin{document}
\title{MACBT: A Multi-Agent Cognitive Behavioral Therapy Decision Support System with Longitudinal Memory}
\author{\IEEEauthorblockN{
De Jiang\textsuperscript{1},
Shuo Zhang\textsuperscript{1},
Weiwei Liao\textsuperscript{2},
Jianying Zhang\textsuperscript{3},\\
Chuanhui Yu\textsuperscript{4},
Hongen Liao\textsuperscript{1},
Kehong Yuan\textsuperscript{1}}
\IEEEauthorblockA{\textsuperscript{1}Tsinghua University, China\\
\texttt{jiangd24tsinghua@gmail.com; zssure@163.com}\\
\texttt{liao@tsinghua.edu.cn; yuankh@sz.tsinghua.edu.cn}\\
\textsuperscript{2}Hangzhou Municipal Health Commission, China; \texttt{3103681@qq.com}\\
\textsuperscript{3}Department of Child Health Care, Hangzhou Women's Hospital\\
(Hangzhou Maternity and Child Health Care Hospital), China\\
\texttt{13758246297@163.com}\\
\textsuperscript{4}Hangzhou Fushu Health Technology Co., Ltd, China\\
\texttt{yuchuanhui2006@163.com}}}
\maketitle

\nocite{r1,r2,r3,r4,r5,r6,r7,r8,r9,r10,r11,r12,r13,r14,r15,r16,r17,r18,r19,r20,r21,r22,r23,r24,r25,r26,r27,r28,r29,r30,r31}

\begin{abstract}
Cognitive behavioral therapy (CBT) is an evidence-based first-line treatment for depression, yet its scale is constrained by the time clinicians spend on pre-session preparation, post-session documentation, and longitudinal cognitive-pathology tracking. We present a clinician-facing AI decision-support system that combines a multi-agent CBT framework (MACBT) with a CBT-specific longitudinal memory module (CD Memory). MACBT encodes the five-stage CBT workflow---assessment, Socratic questioning, cognitive restructuring, behavioral experiments, and treatment monitoring---into five collaborative agents. CD Memory tracks cognitive-distortion type, frequency, severity, and restructuring efficacy across sessions to generate pre-session pathology reports and intervention-priority recommendations. We construct a Chinese CBT dialogue corpus via dual-role large language model simulation and train a Qwen3-14B backbone with supervised fine-tuning and direct preference optimization. Evaluation with GPT-4 judges shows MACBT outperforms MeChat, SoulChat, PsyChat, and CPsyCounX in professionalism (2.62) and clinical authenticity (2.25). The full memory-augmented system further improves session quality by 12.6\% and achieves a longitudinal mean of 2.29 on cross-session continuity, intervention progression, and personalization.
\end{abstract}


\section{Introduction}

Depression is a leading cause of disability worldwide. The World Health Organization estimates that mental disorders affect roughly one in eight people globally \cite{r1}, and the lifetime prevalence of depression in China alone is approximately 6.8\% \cite{r2}. Cognitive behavioral therapy (CBT) is one of the most evidence-based psychotherapies for depression \cite{r3}, \cite{r4}, with meta-analyses showing large effect sizes for mild-to-moderate depression and comparable outcomes to antidepressant medication in many cases. Despite this strong evidence base, trained CBT clinicians are scarce and unevenly distributed \cite{r5}, \cite{r6}; in low-resource settings, the majority of people with depression receive no treatment at all. Expanding access therefore requires raising the productivity and consistency of existing clinicians rather than replacing them.

A complete CBT course typically spans 8-20 sessions. Across this longitudinal process, clinicians face three structural burdens. Pre-session burden: before each appointment, clinicians must review historical notes, assess current cognitive status, and identify unresolved issues from prior interventions. Post-session burden: after each appointment, clinicians manually consolidate session records, assign homework, and track compliance for multiple patients. Longitudinal burden: over the full course, clinicians must monitor the evolution of each patient's cognitive pathology and dynamically adjust the treatment plan. A typical clinician managing 20 active patients may spend 30-45 minutes per patient on pre-session review and post-session documentation alone, compressing the time available for direct therapeutic contact. These burdens limit caseload capacity and increase the risk of information loss between sessions.

Existing digital CBT systems such as Woebot \cite{r7} and Wysa \cite{r8} deliver predefined decision-tree interactions directly to patients. Recent large language model (LLM) systems, including MeChat \cite{r9}, SoulChat \cite{r10}, and CPsyCoun \cite{r11}, improve conversational empathy and professionalism but remain largely patient-facing and single-session. Multi-agent frameworks such as MAGI \cite{r12} structure psychiatric interviews for diagnosis, while generic long-term memory systems such as MemoryBank \cite{r13} lack CBT-specific constructs. None of these works combine CBT treatment logic, multi-agent reasoning, and cross-session cognitive-distortion memory into a clinician-facing decision-support tool.

We address this gap with two integrated contributions. First, we propose MACBT (Multi-Agent Cognitive Behavioral Therapy), which explicitly encodes CBT's five-stage treatment logic into five collaborative agents: an assessment navigator, a Socratic questioning agent, a cognitive restructuring agent, a behavioral experiment agent, and a treatment monitoring agent. Second, we augment MACBT with CD Memory, a longitudinal module that stores the type, frequency, severity, and restructuring efficacy of each patient's cognitive distortions across sessions. CD Memory computes intervention priorities and generates structured pre-session reports, allowing clinicians to move from ``starting from scratch'' to ``intervening with a report.''

We build a Chinese CBT dialogue corpus through dual-role LLM simulation. A patient agent and a counselor agent interact iteratively, with the counselor constrained by the five-stage CBT workflow. The resulting corpus outperforms SMILE, CACTUS, and SimPsyDial on working-alliance quality and CBT technique coverage. We then fully fine-tune a Qwen3-14B backbone \cite{r29} on this corpus and align it with direct preference optimization (DPO) \cite{r28}. Compared with MeChat, SoulChat, PsyChat, and CPsyCounX, MACBT achieves the highest professionalism (2.62) and clinical authenticity (2.25) scores under GPT-4 evaluation. Adding CD Memory raises session quality by 12.6\% and yields a longitudinal mean of 2.29 on cross-session continuity (2.38), intervention progression (2.27), and personalization (2.21).

Our contributions are:
\begin{enumerate}

\item A Chinese CBT dialogue corpus constructed via dual-role LLM simulation, with stronger working-alliance quality and CBT technique coverage than existing open datasets.

\item MACBT, a multi-agent framework that encodes CBT's five-stage logic into collaborative agents and outperforms existing mental-health dialogue systems in professionalism and authenticity.

\item A CBT-specific longitudinal memory module (CD Memory) that tracks cognitive distortions across sessions and supports pre-session planning, substantially improving longitudinal treatment continuity.

\end{enumerate}

\section{Related Work}

\textbf{CBT and digital interventions.} CBT is grounded in Beck's cognitive model, which posits that negative automatic thoughts and cognitive distortions drive emotional distress \cite{r14}. Structured stage models--assessment, intervention, and consolidation--are central to treatment fidelity \cite{r15}. Internet-based CBT (ICBT) has shown effects comparable to face-to-face therapy \cite{r16}, \cite{r17}, and conversational agents such as Woebot \cite{r7} and Wysa \cite{r8} have brought CBT techniques to everyday use. These systems, however, are typically rule-based and patient-facing: they deliver predefined content and cannot adapt to the evolving cognitive state of an individual patient. Clinicians therefore still lack flexible, stage-aware tools that can reason about CBT structure while leaving final decisions in human hands.

\textbf{LLM-based mental-health dialogue.} Recent work has fine-tuned LLMs for emotional support and counseling. SoulChat \cite{r10}, MeChat \cite{r9}, and PsyChat \cite{r19} improve empathy and multi-turn consistency, while CPsyCoun \cite{r11} reconstructs multi-turn dialogues from professional reports. CACTUS \cite{r18} adds a CBT planning agent to guide conversation. These advances show that LLMs can produce clinically flavored text, but the resulting systems remain largely patient-facing dialogue generators. They do not expose the intermediate reasoning--distortion identification, technique selection, progression monitoring--that a clinician needs to trust and supervise an AI assistant.

\textbf{Multi-agent clinical systems.} Multi-agent architectures have been applied to psychiatric assessment. MAGI \cite{r12} uses four collaborative agents to conduct MINI-guided diagnostic interviews. MentalAgora \cite{r20} and MultiAgentESC \cite{r21} explore multi-agent debate and collaboration for emotional support. These works demonstrate the value of agent specialization, but they target diagnosis or general support rather than the sequential treatment stages of CBT. Moreover, they do not maintain a structured, session-crossing record of the patient's cognitive pathology.

\textbf{Long-term memory for dialogue.} Long-term memory mechanisms such as MemoryBank \cite{r13}, generative agents \cite{r22}, and A-MEM \cite{r23} enable personalized, cross-session interactions. In mental health, SOULSPEAK \cite{r24}, MusPsy \cite{r25}, and AnnaAgent \cite{r26} explore session-level memory. However, these memories are generic summaries or seeker simulations; they do not track CBT-specific constructs such as cognitive-distortion type, severity, and restructuring efficacy. A memory module designed for CBT should record not only what was said, but which distortions appeared, how severe they were, and which interventions helped.

\section{Methods}

\subsection{CBT Dialogue Corpus Construction}

To train a model that behaves like a CBT clinician, we first construct a Chinese CBT dialogue corpus through dual-role LLM simulation. We start from 3,134 anonymized psychological counseling reports and extract structured patient personas. The extraction prompt asks an LLM to identify diagnosis, core symptoms, main problem, presenting complaint, emotional triggers, and expected counseling goals. Structuring the persona rather than feeding raw reports into dialogue generation prevents the counselor agent from leaking information the patient would not yet have disclosed.

Each persona drives a patient agent, while a separate counselor agent is constrained by CBT's five-stage workflow: assessment, Socratic questioning, cognitive restructuring, behavioral experiments, and consolidation. The two agents interact iteratively, with the patient agent revealing information gradually rather than possessing the full dialogue history. This design avoids the ``omniscient'' rewriting problem of prior datasets \cite{r9}, \cite{r11}, where a single model knows both sides of the conversation and produces unrealistic therapist-led monologues.

We apply quality constraints to the counselor agent: responses are limited to 40 Chinese characters, avoid numbered lists, contain at most one question, and use conversational rather than didactic language. Dialogues terminate when the patient identifies at least one core distortion, establishes a functional alternative thought, agrees to a behavioral experiment, or expresses readiness to end; otherwise they stop at 50 turns. After filtering incomplete samples, we obtain 2,000 multi-turn dialogues yielding roughly 68,000 supervised fine-tuning samples.

\begin{figure}[t]
\centering
\includegraphics[width=\columnwidth]{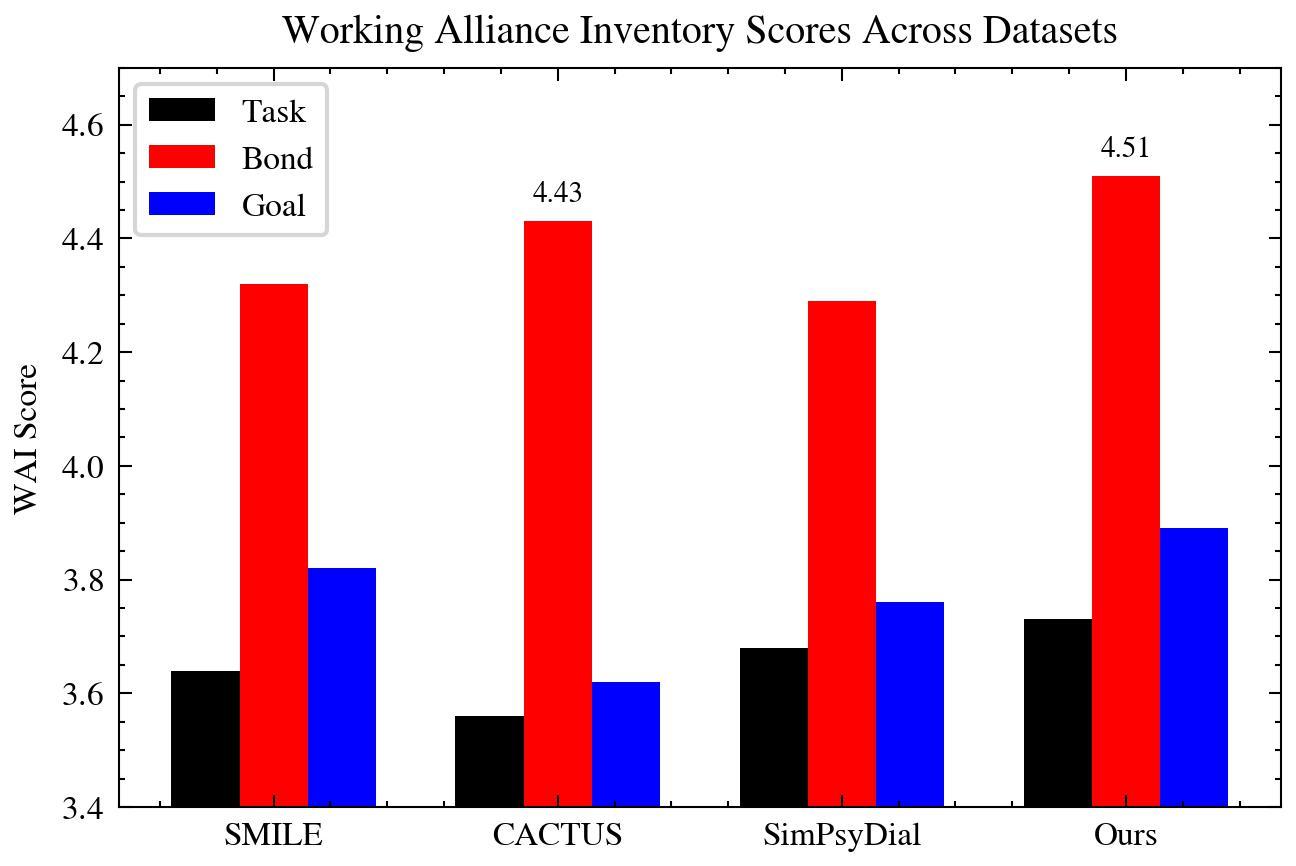}
\caption{Working Alliance Inventory scores across datasets. Our corpus achieves the highest bond (4.51) and goal (3.89) scores.}
\label{fig:1}
\end{figure}

We evaluate the corpus on patient consistency, counselor professionalism, emotional realism, and CBT technique coverage. Patient consistency is measured by lexical overlap (0.386) and semantic similarity (0.801) between patient utterances and assigned personas, both substantially higher than a random-patient baseline. Counselor professionalism is assessed with the Working Alliance Inventory (WAI) \cite{r27}: our corpus scores 3.89 on goal, 3.73 on task, and 4.51 on bond, outperforming SMILE, CACTUS, and SimPsyDial (Fig.~\ref{fig:1}).

Emotional realism is measured by emotional variability, emotional-word density, and contextual responsiveness (Table~\ref{tab:I}). Our corpus scores 0.287, 0.171, and 0.421, respectively, approaching the RealCBT benchmark (0.312, 0.184, 0.463) and substantially exceeding the simulated baselines. \begin{table}[t]
\caption{Emotional-arc comparison across datasets.}
\label{tab:I}
\centering
\footnotesize
\setlength{\tabcolsep}{3pt}
\begin{tabular}{lccc}
\hline
Dataset & Variability & Word density & Responsiveness \\
\hline
RealCBT & 0.312 & 0.184 & 0.463 \\
SMILE & 0.198 & 0.132 & 0.289 \\
CACTUS & 0.241 & 0.149 & 0.347 \\
SimPsyDial & 0.183 & 0.128 & 0.274 \\
Ours & 0.287 & 0.171 & 0.421 \\
\hline
\end{tabular}
\end{table}

The improvement is largest on contextual responsiveness (0.421 vs. 0.289 for SMILE and 0.274 for SimPsyDial), indicating that the counselor agent adapts its questioning strategy to the patient's immediate emotional signals rather than following a fixed script. The emotional-variability gap between our corpus and RealCBT (0.287 vs. 0.312) is only 0.025, whereas SMILE and SimPsyDial lag by more than 0.11. This pattern arises because the patient agent does not possess the full dialogue history: it must react spontaneously from its own cognitive perspective, producing genuine emotional exploration rather than the flat, overly cooperative tone typical of omniscient-rewriting datasets.

\begin{figure}[t]
\centering
\includegraphics[width=\columnwidth]{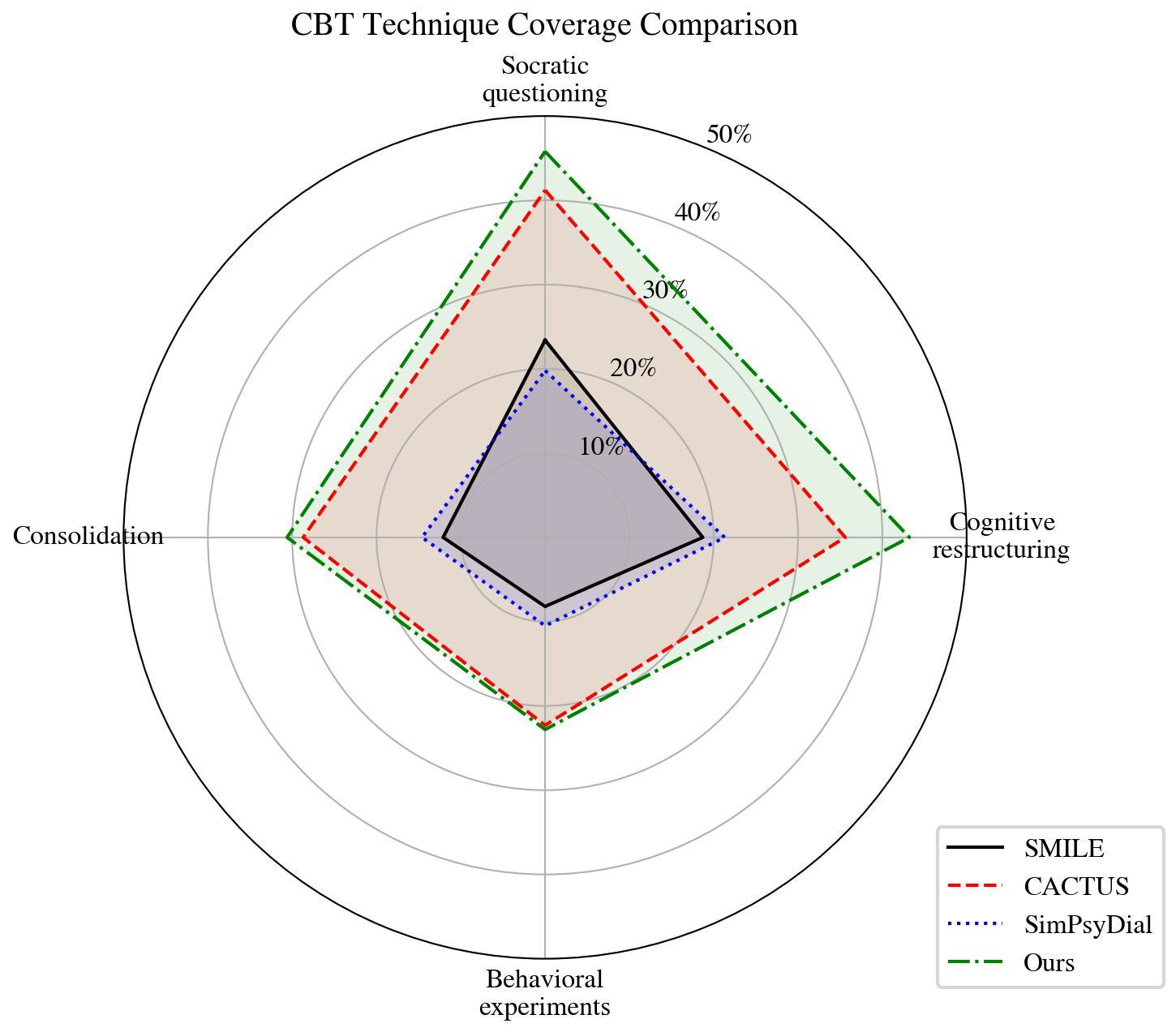}
\caption{CBT technique coverage comparison across SMILE, CACTUS, SimPsyDial, and our corpus. Our corpus leads on all four CBT techniques and achieves the highest overall coverage.}
\label{fig:2}
\end{figure}

\begin{figure}[t]
\centering
\includegraphics[width=\columnwidth]{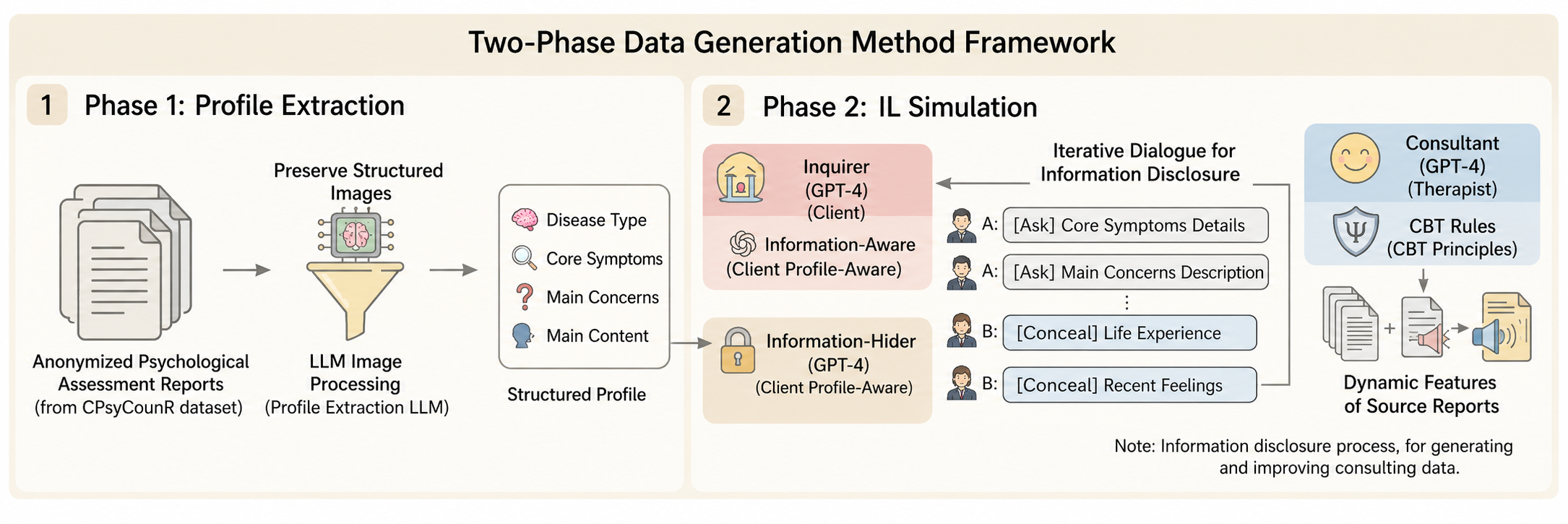}
\caption{Two-stage CBT dialogue corpus construction. Stage 1 extracts structured patient personas from anonymized counseling reports; Stage 2 simulates multi-turn dialogues between a profile-aware patient agent and a CBT-principled counselor agent.}
\label{fig:3}
\end{figure}

Fig.~\ref{fig:3} summarizes the two-stage corpus construction pipeline: structured persona extraction followed by iterative dual-role LLM simulation. Fig.~\ref{fig:2} compares CBT technique coverage across datasets. Our corpus leads on Socratic questioning (45.8\%), cognitive restructuring (43.2\%), behavioral experiments (22.8\%), and consolidation (30.6\%), with an overall coverage of 35.6\% versus 31.9\% for the strongest baseline. The improvement is largest on cognitive restructuring, the central CBT intervention, indicating that the counselor agent successfully internalized stage-appropriate techniques.

\subsection{MACBT Framework}

MACBT is a multi-agent framework that explicitly models the CBT treatment workflow. A Qwen3-14B backbone \cite{r29} is fully fine-tuned on the CBT corpus for two epochs (learning rate $10^{-5}$, cosine schedule, 16-bit precision) and then aligned with DPO \cite{r28} using 1,347 preference pairs. The DPO pairs are constructed by sampling four candidate responses per context, scoring them along comprehensiveness, professionalism, authenticity, and safety, and selecting the highest- and lowest-scoring replies.

\begin{figure}[t]
\centering
\includegraphics[width=\columnwidth]{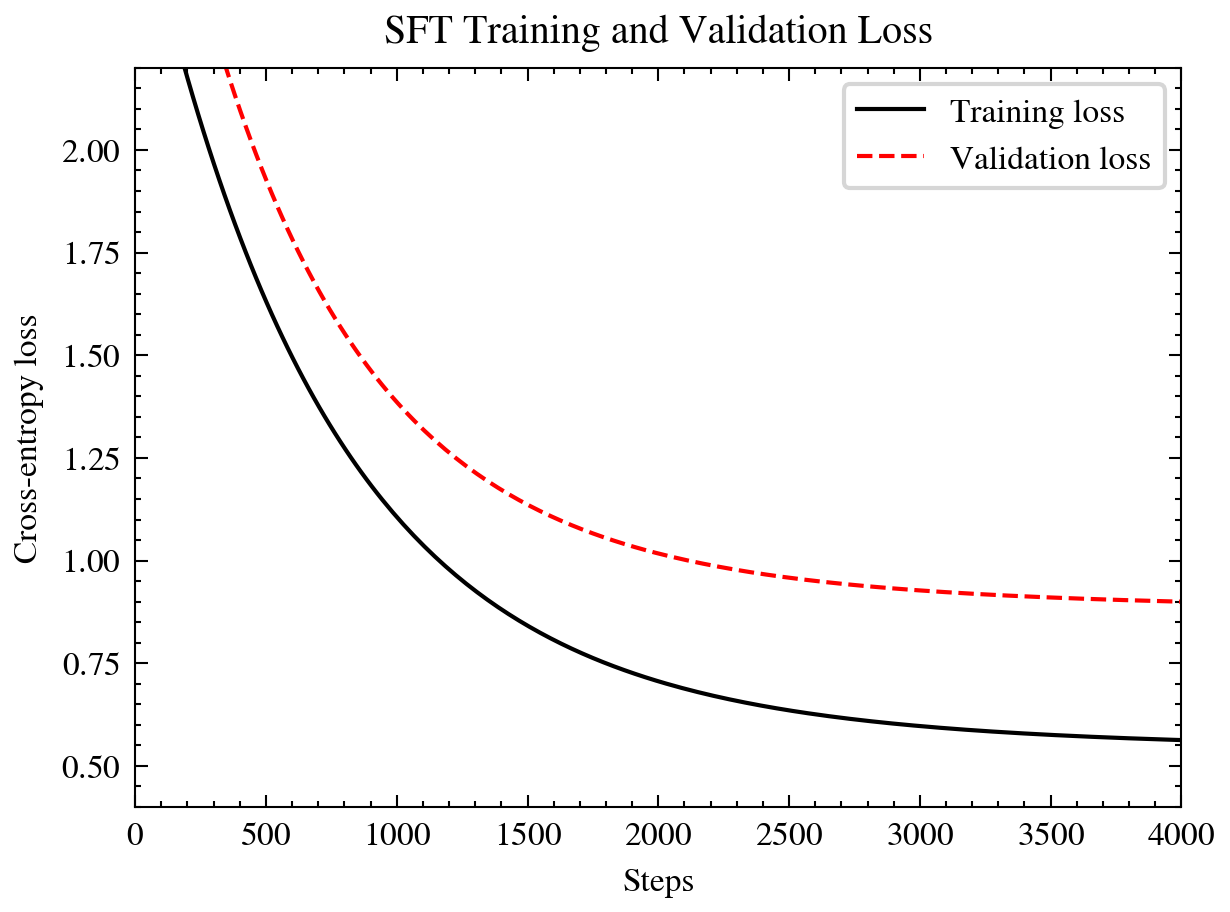}
\caption{SFT training and validation cross-entropy loss over 4,000 steps. The final training loss is 0.536 and the final validation loss is 0.880.}
\label{fig:4}
\end{figure}

\begin{figure}[t]
\centering
\includegraphics[width=\columnwidth]{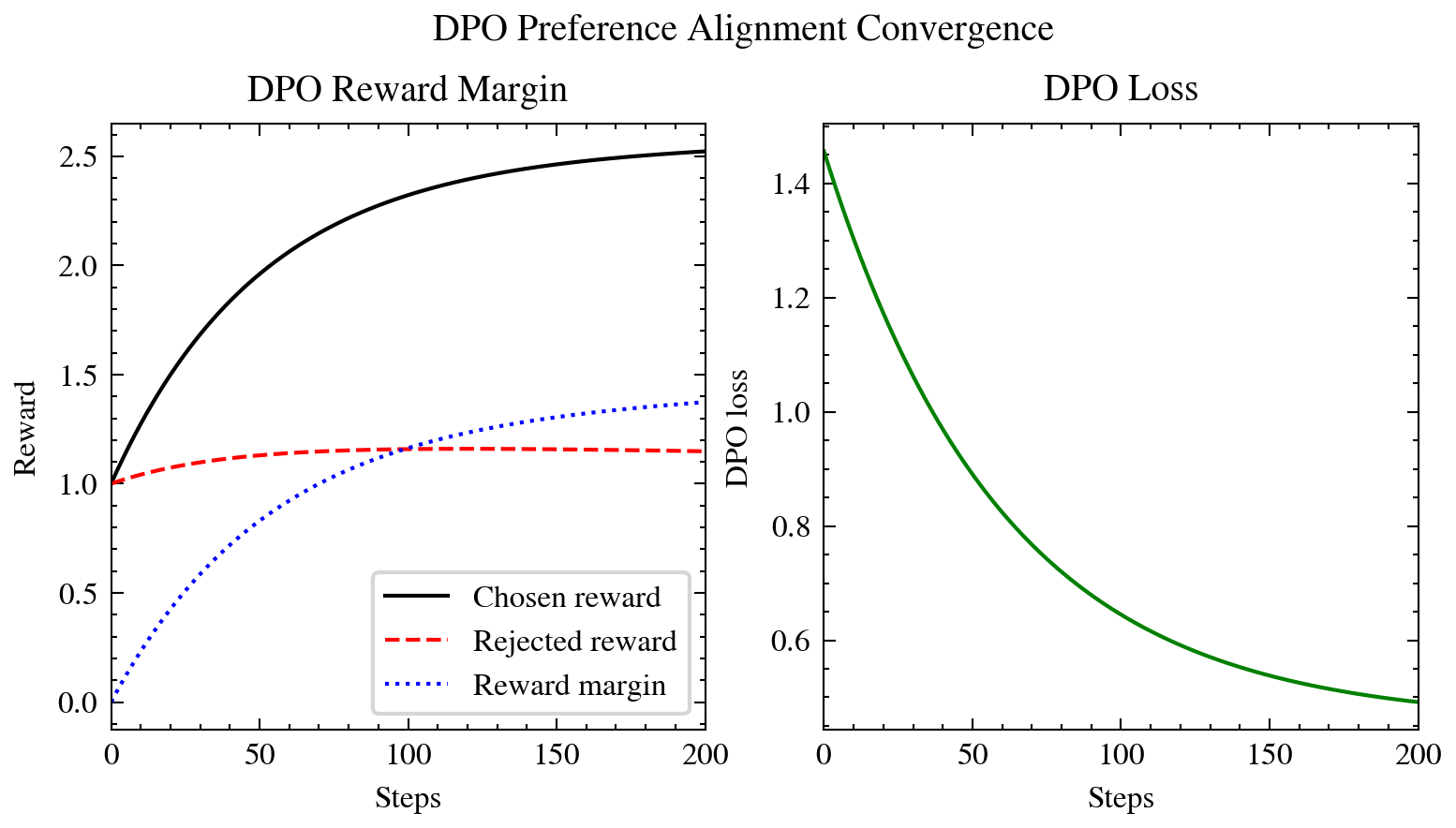}
\caption{DPO preference alignment over 200 steps. The reward gap between chosen and rejected responses widens, the reward margin reaches 2.07, and the DPO loss converges to 0.456.}
\label{fig:5}
\end{figure}

Fig.~\ref{fig:4} shows the SFT training and validation loss curves, and Fig.~\ref{fig:5} shows DPO reward margins and DPO loss convergence.

Table~\ref{tab:II} lists the training configuration. All experiments use 16-bit precision and a cosine learning-rate schedule with a short warmup.

\begin{table}[t]
\caption{Training hyperparameters.}
\label{tab:II}
\centering
\footnotesize
\setlength{\tabcolsep}{3pt}
\begin{tabular}{ll}
\hline
Hyperparameter & Setting \\
\hline
Backbone & Qwen3-14B \\
GPU & 4 $\times$ NVIDIA A100 80G \\
Per-device batch size & 1 \\
Gradient accumulation steps & 2 \\
Learning rate & $10^{-5}$ \\
LR scheduler & cosine annealing \\
Training epochs & 2 \\
Precision & 16-bit \\
DPO preference pairs & 1,347 \\
\hline
\end{tabular}
\end{table}

The five agents share a working memory and operate under the assessment navigator's scheduling:

\begin{itemize}
\item \textbf{Assessment navigator} maintains the CBT stage structure and decides when to transition between assessment, questioning, restructuring, experiments, and consolidation.
\item \textbf{Socratic questioning agent} generates clarification, hypothesis-testing, evidence-testing, and perspective-taking questions to help patients discover distortions.
\item \textbf{Cognitive restructuring agent} identifies ten categories of cognitive distortion and applies appropriate techniques (e.g., continuum for all-or-nothing thinking, counterexamples for overgeneralization, probability estimation for catastrophizing).
\item \textbf{Behavioral experiment agent} designs SMART behavioral experiments and tracks execution.
\item \textbf{Treatment monitoring agent} evaluates session progress and produces a structured report for the clinician.
\end{itemize}

\begin{figure}[t]
\centering
\includegraphics[width=\columnwidth]{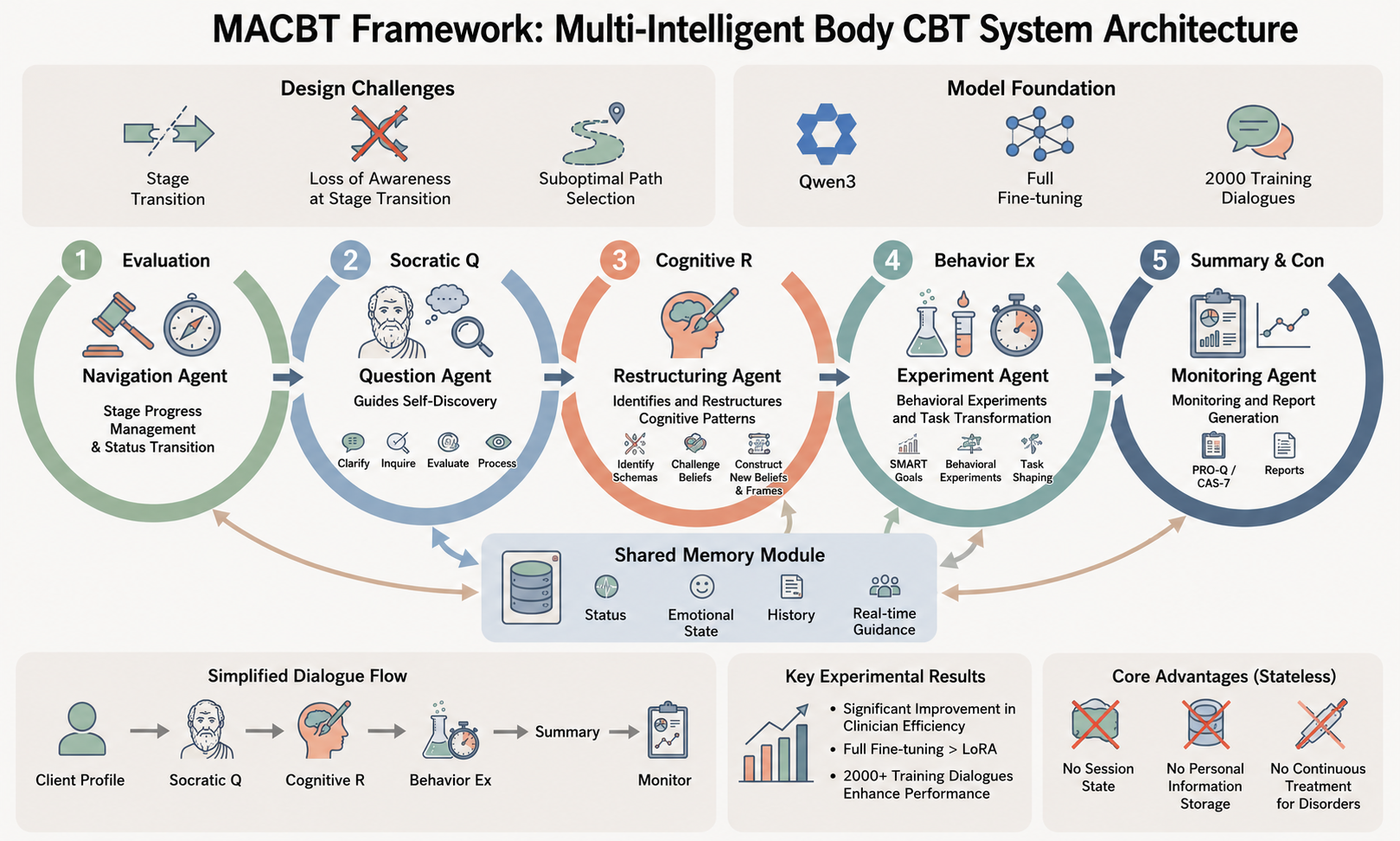}
\caption{Overview of MACBT. The assessment navigator schedules one of four intervention agents; all agents read from and write to shared working memory, and the treatment monitoring agent outputs a structured clinician report.}
\label{fig:6}
\end{figure}

Fig.~\ref{fig:6} illustrates the framework. The shared working memory stores the current stage, identified distortions, and intervention history so that each agent's decision is grounded in the evolving session state.

\begin{figure}[t]
\centering
\includegraphics[width=\columnwidth]{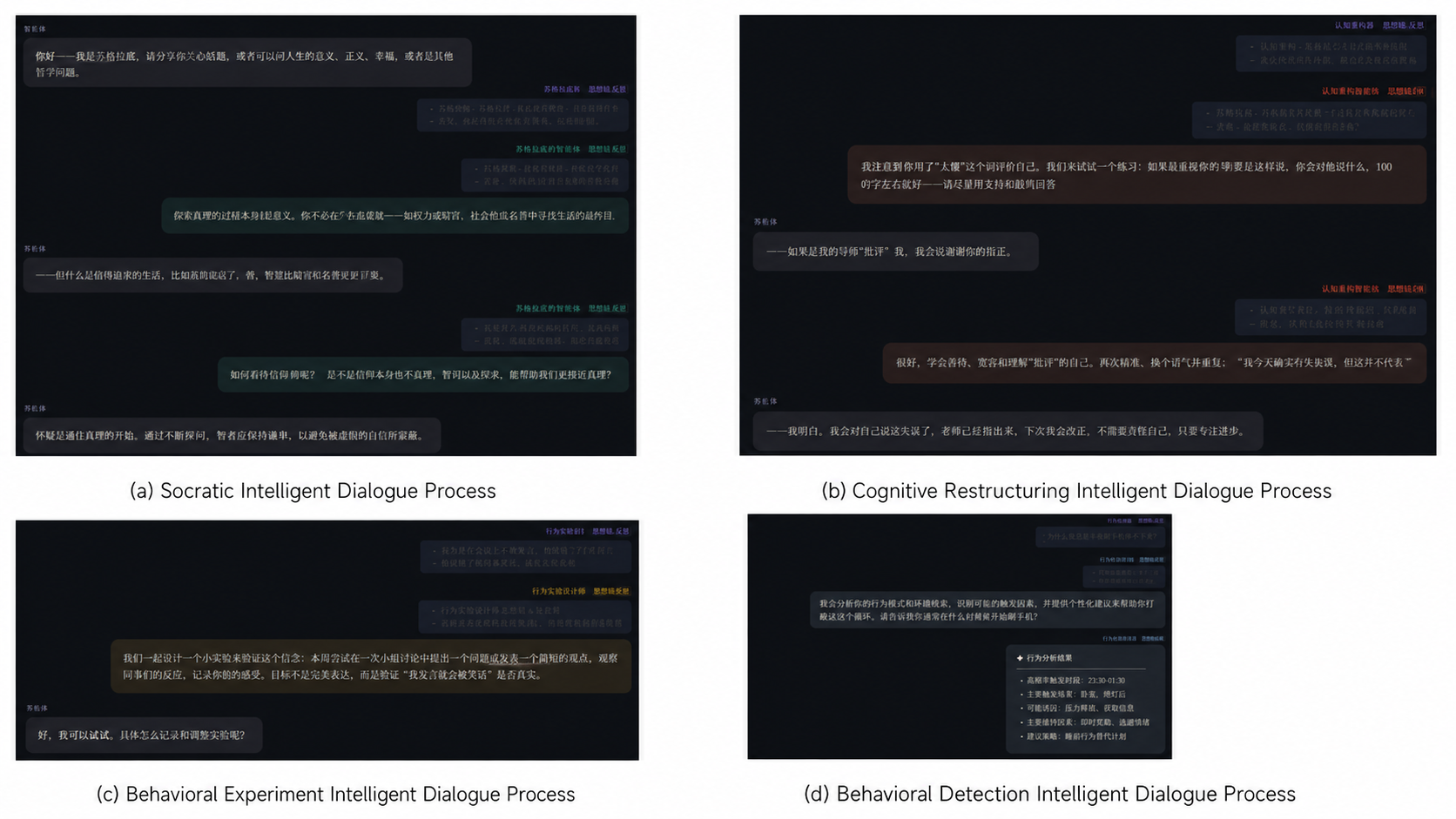}
\caption{Representative MACBT session excerpts: (a) Socratic questioning, (b) cognitive restructuring, (c) behavioral experiment, and (d) treatment monitoring.}
\label{fig:7}
\end{figure}

\textbf{Example session.} Fig.~\ref{fig:7} shows a representative excerpt from a MACBT session. After the assessment navigator identifies an ``all-or-nothing'' distortion around exam failure, the Socratic questioning agent probes the evidence, the cognitive restructuring agent invites a continuum rating, the behavioral experiment agent proposes a small behavioral task, and the treatment monitoring agent records the distortion and intervention efficacy for the next session.

\subsection{Longitudinal Memory System}

Standard MACBT is stateless: each session starts without knowledge of prior distortions or interventions. We add a dual-loop longitudinal memory system consisting of an intra-session loop and an inter-session loop, connected through a CBT-specific cognitive-distortion memory (CD Memory).

Basic Memory extracts high-level personal and psychological insights from each patient utterance and stores them as structured summaries. Rather than retaining raw dialogue history, Basic Memory keeps distilled insight entries--core beliefs, emotional triggers, significant life events, and behavioral patterns--that are semantically indexed for retrieval. This design reduces context-window pressure and ensures that the dynamic prompt contains only information relevant to the current intervention target. After each session, long-term clinically valuable insights are merged into a persistent patient profile, enabling gradual deepening of the system's understanding across the treatment course.

CD Memory is the core longitudinal store. For each detected cognitive distortion, it records the distortion type, the triggering utterance, a 1-5 severity score, and the history of restructuring efficacy. Efficacy is scored from the patient's immediate response after each intervention (attitude shift, emotional improvement, and willingness to try a behavioral experiment).

At the start of each session, CD Memory computes an intervention priority score for each distortion type by combining recency, frequency, and severity. Let r, f, and s denote normalized recency, frequency, and severity, and let $\alpha$, $\beta$, and $\gamma$ be their weights. The priority score is\begin{equation}P=\alpha r+\beta f+\gamma s.\label{eq:priority}\end{equation} The distortion with the highest $P$ becomes the current session's primary intervention target. The system also retrieves historically effective and ineffective restructuring techniques for that distortion from the efficacy record. Efficacy scores are stored as structured tuples (distortion type, technique, efficacy score, session id), enabling the system to build a patient-specific technique-response map over time. For example, if continuum techniques consistently yield high efficacy for a patient's all-or-nothing thinking while probability estimation performs poorly, the dynamic prompt will recommend continuum and avoid probability estimation for that distortion type in subsequent sessions.

\textbf{Dynamic prompt generation.} The static prompt defines the CBT role and safety constraints; the dynamic prompt is assembled each turn from (1) the priority distortion and its background, (2) recommended and avoided restructuring techniques, and (3) the current stage from the CBT usage log.

\begin{figure}[t]
\centering
\includegraphics[width=\columnwidth]{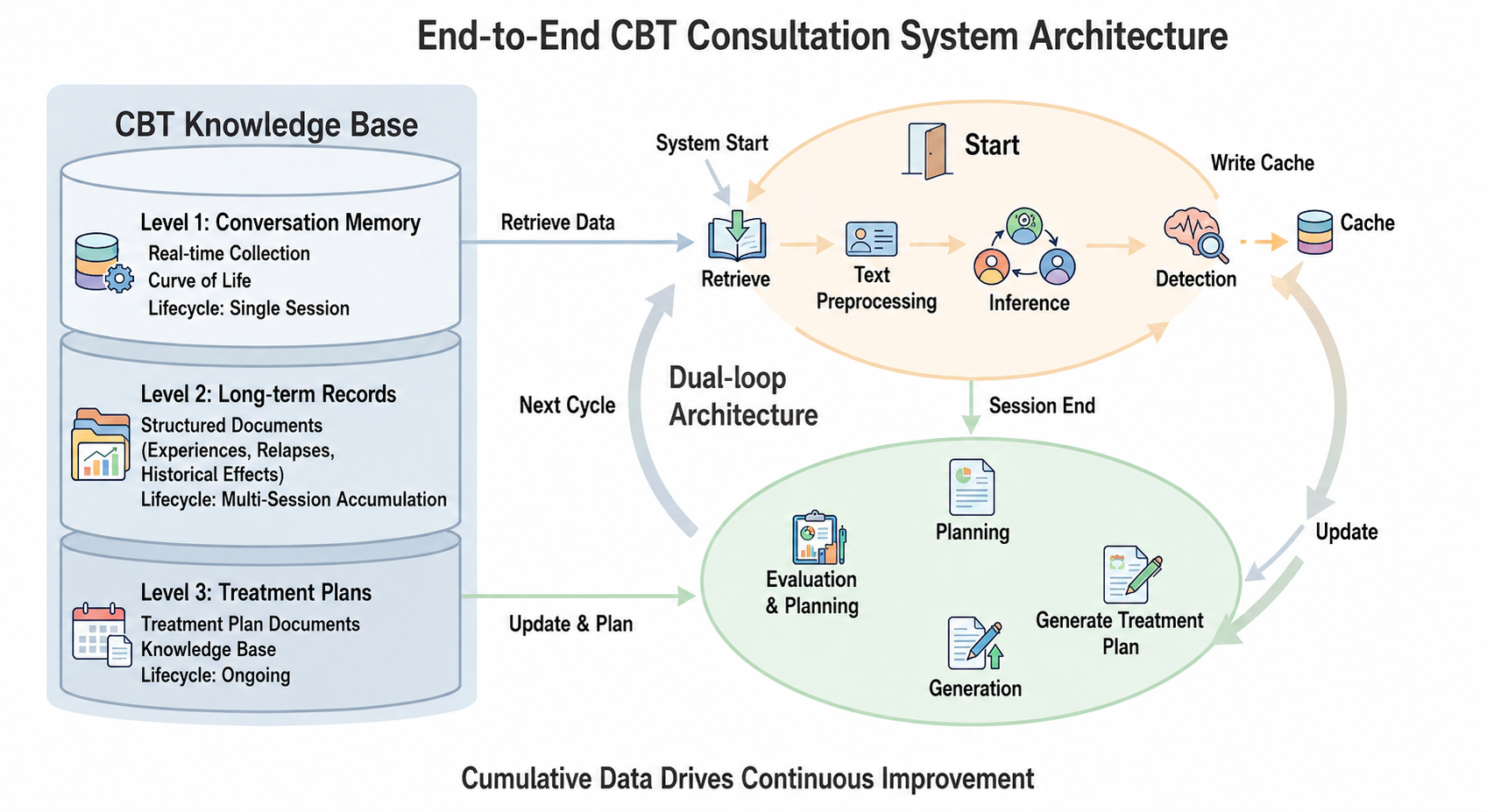}
\caption{Dual-loop longitudinal memory architecture. The intra-session loop uses Basic Memory and CD Memory to guide live dialogue; the inter-session loop evaluates the session and updates CD Memory and the treatment plan.}
\label{fig:8}
\end{figure}

\textbf{Inter-session loop.} After each session, the treatment monitoring agent evaluates cognitive-distortion improvement, restructuring-technique efficacy, and behavioral-experiment completion. These assessments update CD Memory and generate a structured treatment plan for the next session, which is injected into the next session's dynamic prompt. Fig.~\ref{fig:8} shows the dual-loop architecture.

\section{Experiments and Results}

\subsection{Evaluation Setup}

We evaluate MACBT and the memory-augmented system with GPT-4 as a judge \cite{r30}, \cite{r31}. Session quality is scored on four 0-3 Likert dimensions: comprehensiveness, professionalism, authenticity, and safety. Longitudinal quality is scored on three 0-3 dimensions: cross-session continuity, intervention progression, and personalization. We use 100 virtual patient profiles for the single-session evaluation and 30 three-session test units for the longitudinal evaluation.

\begin{table}[t]
\caption{Session-quality comparison (0--3 scale, higher is better).}
\label{tab:III}
\centering
\footnotesize
\setlength{\tabcolsep}{3pt}
\begin{tabular}{lcccc}
\hline
System & Comp. & Prof. & Auth. & Mean \\
\hline
MeChat \cite{r9} & 1.28 & 1.95 & 1.88 & 1.46 \\
SoulChat \cite{r10} & 1.35 & 2.01 & 2.12 & 1.54 \\
PsyChat \cite{r19} & 1.42 & 2.23 & 2.08 & 1.62 \\
CPsyCoun \cite{r11} & 1.56 & 2.48 & 2.11 & 1.71 \\
MACBT & 1.44 & 2.62 & 2.25 & 1.83 \\
\hline
\end{tabular}
\end{table}

\subsection{Main Results}

Table~\ref{tab:III} compares MACBT against MeChat, SoulChat, PsyChat, and CPsyCounX. MACBT achieves the highest professionalism (2.62) and authenticity (2.25) and the best overall mean (1.83). Its comprehensiveness is slightly lower than CPsyCounX because MACBT's navigator focuses depth-first on the current CBT stage rather than listing multiple topics.

\textbf{Analysis.} The professionalism lead of MACBT (2.62) over CPsyCounX (2.48) reflects two advantages: CBT-specific fine-tuning embeds structured Socratic questioning and cognitive restructuring into the model's generation patterns, and the navigator's hard stage constraints prevent treatment-phase skipping. The authenticity lead (2.25 vs. 2.11) reflects the dual-role corpus design: by simulating information asymmetry between patient and counselor, the training data captures exploratory, responsive dialogue rather than the flat, directive tone of omniscient-rewriting approaches.

Table~\ref{tab:IV} shows the impact of fine-tuning and preference alignment. A non-fine-tuned Qwen3-14B with the MACBT prompt scores only 1.74 in professionalism; generic mental-health data raises it to 2.08, while CBT-specific data reaches 2.48. Adding DPO improves professionalism by 0.14 and authenticity by 0.27.

\textbf{Fine-tuning strategy.} Table~\ref{tab:V} compares parameter-efficient LoRA adapters against full fine-tuning. LoRA with rank 16 underfits the CBT task (professionalism 2.21). Increasing rank to 64 narrows the gap (professionalism 2.38) but still trails full fine-tuning on professionalism and mean score. Full fine-tuning is therefore used for the final model.

\begin{table}[t]
\caption{Ablation on training strategy (0--3 scale).}
\label{tab:IV}
\centering
\footnotesize
\setlength{\tabcolsep}{3pt}
\begin{tabular}{lccc}
\hline
Setting & Prof. & Auth. & Mean \\
\hline
No fine-tuning & 1.74 & 1.98 & 1.55 \\
Generic mental-health SFT & 2.08 & 2.11 & 1.68 \\
CBT-specific SFT & 2.48 & 2.12 & 1.92 \\
CBT SFT + DPO & 2.62 & 2.25 & 2.10 \\
\hline
\end{tabular}
\end{table}

\begin{table}[t]
\caption{Comparison of fine-tuning strategies (0--3 scale).}
\label{tab:V}
\centering
\footnotesize
\setlength{\tabcolsep}{3pt}
\begin{tabular}{lcccc}
\hline
Strategy & Comp. & Prof. & Auth. & Mean \\
\hline
LoRA (rank=16) & 1.18 & 2.21 & 1.97 & 1.79 \\
LoRA (rank=64) & 1.31 & 2.38 & 2.09 & 1.93 \\
Full fine-tuning & 1.31 & 2.48 & 1.98 & 1.92 \\
\hline
\end{tabular}
\end{table}

\begin{figure}[t]
\centering
\includegraphics[width=\columnwidth]{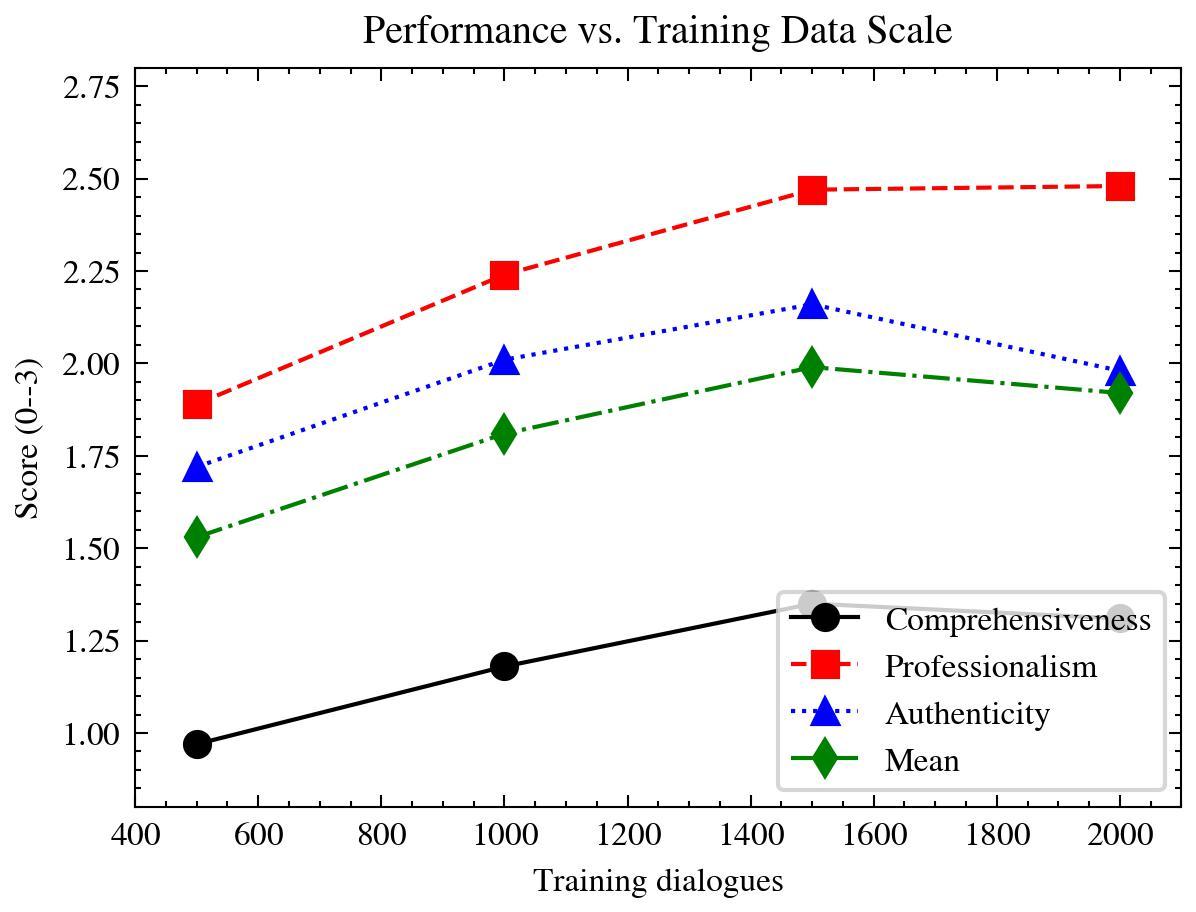}
\caption{Performance versus training data scale. Professionalism is the most responsive dimension, rising by 0.73 from 500 to 2,000 dialogues.}
\label{fig:9}
\end{figure}

\textbf{Training data scale.} Table~\ref{tab:VI} and Fig.~\ref{fig:9} report SFT performance as the number of training dialogues increases from 500 to 2,000. Professionalism improves monotonically (1.89 to 2.48), while comprehensiveness and authenticity saturate after 1,500 dialogues. The gap between 1,500 and 2,000 dialogues still yields a 0.15 professionalism gain, suggesting larger CBT-specific corpora remain valuable. DPO then lifts the final professionalism to 2.62.

\begin{table}[t]
\caption{Ablation on training data scale (0--3 scale).}
\label{tab:VI}
\centering
\footnotesize
\setlength{\tabcolsep}{3pt}
\begin{tabular}{lcccc}
\hline
\# Dialogues & Comp. & Prof. & Auth. & Mean \\
\hline
500 & 0.97 & 1.89 & 1.72 & 1.53 \\
1,000 & 1.18 & 2.24 & 2.01 & 1.81 \\
1,500 & 1.35 & 2.47 & 2.16 & 1.99 \\
2,000 & 1.31 & 2.48 & 1.98 & 1.92 \\
\hline
\end{tabular}
\end{table}

\subsection{Longitudinal Results}

Table~\ref{tab:VII} compares the full memory-augmented system against MACBT without memory, MACBT with a generic cross-session summary, and MACBT with CD Memory limited to a single session. The full system achieves the highest session quality (2.06) and longitudinal mean (2.29). The large gap between the single-session CD Memory variant (1.72) and the full system (2.29) confirms that cross-session persistence is essential for longitudinal continuity. Notably, the generic cross-session summary improves continuity (0.87 to 1.48) but falls far short of the structured CD Memory, because generic summaries lose the distortion-level granularity that drives CBT intervention planning.

The longitudinal mean of 2.29 reflects three clinically meaningful improvements: cross-session continuity (2.38) prevents repeated rediscovery of the same issues, intervention progression (2.27) confirms appropriate treatment focus adjustment, and personalization (2.21) reflects technique selection tailored to accumulated efficacy records.

\begin{figure}[t]
\centering
\includegraphics[width=\columnwidth]{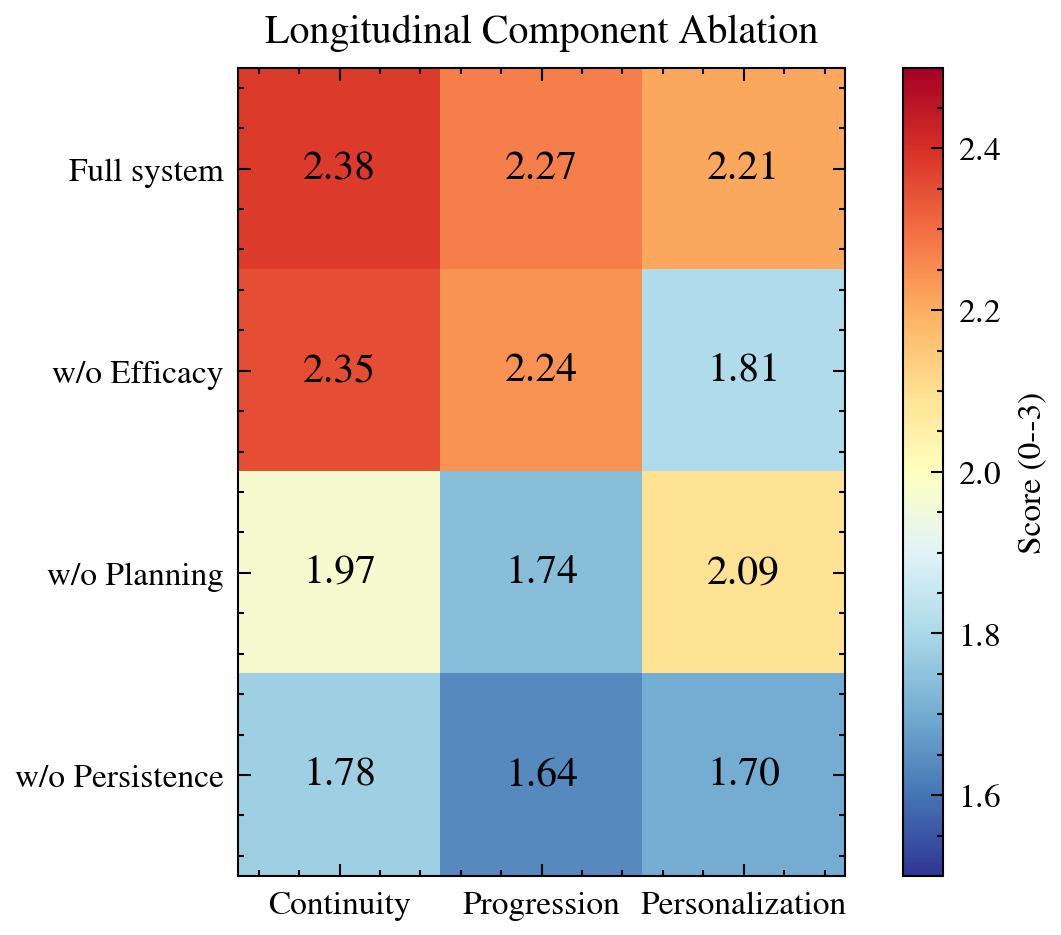}
\caption{Heatmap of longitudinal ablation results. Removing cross-session persistence causes the largest drop across all three longitudinal dimensions.}
\label{fig:10}
\end{figure}

Table~\ref{tab:VIII} and Fig.~\ref{fig:10} ablate individual components of the longitudinal system. Removing restructuring-efficacy records primarily hurts personalization (2.21 to 1.81), because the system can no longer match distortions to historically effective techniques. Removing inter-session planning most strongly reduces intervention progression (2.27 to 1.74), since each session must then restart without a staged plan. Removing cross-session persistence degrades all three metrics, confirming that memory must survive across sessions to yield continuity.

\begin{table}[t]
\caption{Longitudinal comparison over three sessions (0--3 scale).}
\label{tab:VII}
\centering
\footnotesize
\setlength{\tabcolsep}{3pt}
\begin{tabular}{lcccc}
\hline
System & Session quality & Cont. & Prog. & Long. mean \\
\hline
MACBT (no memory) & 1.83 & 0.91 & 0.89 & 0.87 \\
+ Generic summary & 1.89 & 1.52 & 1.47 & 1.48 \\
+ Single-session CD & 1.96 & 1.78 & 1.71 & 1.72 \\
Full CD Memory & 2.06 & 2.38 & 2.27 & 2.29 \\
\hline
\end{tabular}
\end{table}

\begin{table}[t]
\caption{Ablation of longitudinal memory components (0--3 scale).}
\label{tab:VIII}
\centering
\footnotesize
\setlength{\tabcolsep}{3pt}
\begin{tabular}{lcccc}
\hline
System variant & Cont. & Prog. & Pers. & Mean \\
\hline
Full CD Memory & 2.38 & 2.27 & 2.21 & 2.29 \\
-- Efficacy records & 2.35 & 2.24 & 1.81 & 2.13 \\
-- Inter-session planning & 1.97 & 1.74 & 2.09 & 1.93 \\
-- Cross-session persistence & 1.78 & 1.64 & 1.70 & 1.71 \\
\hline
\end{tabular}
\end{table}

\section{Discussion and Conclusion}

We presented MACBT, a multi-agent decision-support system for depression CBT, and CD Memory, a CBT-specific longitudinal memory module. The system reduces three clinician burdens: pre-session preparation through a CBT dialogue corpus and pre-session reports, in-session decision support through structured multi-agent reasoning, and longitudinal tracking through cross-session cognitive-distortion memory.

\subsection{Key Findings}

Three results stand out. First, encoding CBT stage logic explicitly into specialized agents improves professionalism and clinical authenticity compared with general mental-health dialogue systems. The gap over generic mental-health SFT (2.08 vs. 2.48 professionalism) shows that domain-specific data matter, and the further gain from DPO (+0.14 professionalism, +0.27 authenticity) shows that preference alignment can refine clinical style. The improvement is largest on authenticity, indicating that DPO specifically suppresses generic, overly agreeable responses in favor of clinically appropriate ones.

Second, full fine-tuning outperforms LoRA adapters in this setting, suggesting that the multi-agent prompts and stage transitions benefit from deeper parameter updates than a rank-64 adapter can provide. The rank-64 LoRA reaches 2.38 professionalism, close to full fine-tuning's 2.48, but its mean score still trails, possibly because LoRA cannot fully adapt the model's latent representation of CBT stage boundaries.

Third, the longitudinal module is the largest single contributor to cross-session continuity. Even a single-session CD Memory raises the longitudinal mean from 0.87 to 1.72, and full cross-session persistence reaches 2.29. This gain is larger than the gain from switching from generic to CBT-specific SFT, suggesting that for longitudinal CBT, structured memory may be as important as high-quality dialogue generation. From a clinical perspective, the pre-session report generated by CD Memory transforms the clinician's workflow from reconstructing patient history from scattered notes to reviewing a structured, distortion-level summary with intervention recommendations. This shift directly addresses the pre-session burden identified in the introduction and provides a concrete path toward scalable CBT delivery.

\subsection{Limitations}

Limitations remain. The evaluation relies on GPT-4 judges and simulated patients, not licensed clinicians or real patients. Although GPT-4 judges correlate reasonably with human ratings on open-ended dialogue evaluation \cite{r31}, they may overweight fluent surface form and under-weight clinical nuance. The longitudinal test covers only three sessions, far shorter than a standard 8-20 session CBT course. We therefore cannot claim that the memory design scales to the full vocabulary of distortions and techniques accumulated over months of therapy.

Data construction also carries constraints. The corpus is derived from anonymized counseling reports rather than verbatim CBT transcripts, so the simulated dialogues approximate rather than reproduce real therapist--patient interaction. Patient personas are static, whereas real patients evolve in response to treatment. In addition, the 2,000-dialogue corpus, while sufficient to demonstrate the framework, is modest by modern LLM training standards; the performance curve suggests that further gains are likely with larger corpora. Safety guardrails, crisis detection, and regulatory compliance require further work before clinical deployment.

\subsection{Ethical Considerations and Safety}

Because MACBT targets a vulnerable population and a high-stakes clinical domain, safety is a design requirement rather than an afterthought. The current system includes three layers of protection. At the content layer, the counselor agent prompt explicitly forbids medical diagnosis, medication advice, and crisis normalization; it instead routes urgent language to a predefined escalation message. At the workflow layer, the assessment navigator monitors session stage and can terminate or hand off when the dialogue drifts outside the supported CBT scope. At the human oversight layer, all outputs are framed as draft suggestions for a clinician, not as direct patient advice.

Several risks remain unresolved. LLM judges may be influenced by response length and fluency, so human clinician validation is essential before any real-world use. The corpus is built from Chinese counseling reports and may not transfer to other languages or cultural contexts without retraining. Bias in the original reports could propagate through the simulation into model behavior. Finally, memory modules raise privacy concerns: longitudinal records of cognitive distortions are sensitive and must be stored with strict access controls, encryption, and retention limits. Any deployment should comply with local mental-health and data-protection regulations.

\subsection{Future Work}

Future directions include clinician-in-the-loop validation over a full 8-20 session treatment course, memory compression for extended therapy courses, multimodal cues to enrich assessment decisions, and learning CD Memory priority weights from clinician feedback. The CD Memory design could also be adapted to other structured psychotherapies such as Dialectical Behavior Therapy (DBT) or Acceptance and Commitment Therapy (ACT).

\bibliographystyle{IEEEtran}
\bibliography{references}
\end{document}